\documentclass[conference]{IEEEtran}
\IEEEoverridecommandlockouts
\usepackage{cite}
\usepackage{amsmath,amssymb,amsfonts}
\usepackage{graphicx}

\usepackage{subcaption}
\usepackage{multirow}
\usepackage{pifont}
\usepackage{diagbox}
\usepackage[dvipsnames]{xcolor}
\usepackage{booktabs}
\usepackage{hyperref}
\usepackage{caption}

\newcommand{\cmark}{\ding{51}}%
\newcommand{\xmark}{\ding{55}}%
\newcommand{\etal}{\textit{et al.}}

\def\BibTeX{{\rm B\kern-.05em{\sc i\kern-.025em b}\kern-.08em
    T\kern-.1667em\lower.7ex\hbox{E}\kern-.125emX}}
\begin{document}

\title{CLIP-EBC: CLIP Can Count Accurately through Enhanced Blockwise Classification}

\author{
Yiming Ma \\
Department of Computer Science \\
University of Warwick \\
Coventry, United Kingdom \\
{\tt\small yiming.ma.1@warwick.ac.uk}
\and
Victor Sanchez \\
Department of Computer Science \\
University of Warwick \\
Coventry, United Kingdom \\
{\tt\small V.F.Sanchez-Silva@warwick.ac.uk}
\and
Tanaya Guha \\
School of Computing Science \\
University of Glasgow \\
Glasgow, United Kingdom \\
{\tt\small Tanaya.Guha@glasgow.ac.uk}
}

\maketitle

\begin{abstract}
We propose \textbf{CLIP-EBC}, the \textit{first fully CLIP-based model designed for accurate crowd density estimation}.
Existing classification-based crowd counting frameworks face challenges when directly applied with CLIP. 
For instance, these methods quantize count values into bordering real-valued bins, which are inconsistent with CLIP's pretraining corpus.
Besides, this quantization strategy also introduces label ambiguity near shared bin borders.
To address these issues, we propose the \textbf{E}nhanced \textbf{B}lockwise \textbf{C}lassification (\textbf{EBC}) framework, which utilizes integer-valued bins and additionally incorporates a density-map-based loss to further improve count accuracy.
Building on this backbone-agnostic framework, \textbf{CLIP-EBC} fully leverages CLIP’s recognition capabilities for crowd counting.
Experiments show that EBC improves classification-based methods by up to 44.5\% in mean absolute error (MAE) on the UCF-QNRF dataset.
Furthermore, CLIP-EBC achieves state-of-the-art performance on the NWPU-Crowd test set, with an MAE of 58.2, surpassing the previous best method by 8.6\%.
Our code is available at {\url{https://github.com/Yiming-M/CLIP-EBC}}.
\end{abstract}
\vspace{-8pt}
\section{Introduction}
\label{sec:intro}
%
%
The vision-language model CLIP \cite{radford2021learning} is pre-trained on millions of image-text pairs using contrastive learning, enabling it to align visual and textual modalities effectively.
This capability has allowed CLIP to excel in pattern recognition, setting new benchmarks in downstream tasks such as zero-shot image classification \cite{radford2021learning}, 
and semantic segmentation \cite{luddecke2022image}.
Despite its strong performance in these recognition-based tasks, its potential for regression tasks, which require predicting continuous values, remains largely unexplored.
This limitation arises primarily from its contrastive learning objective, which is inherently designed for discrete decision-making tasks, such as classification or retrieval, where the goal is to match or rank embeddings.
However, regression tasks require predicting continuous values, which is not the focus of contrastive learning.

In this work, we aim to address this gap by investigating whether CLIP can effectively perform regression tasks, with a specific focus on crowd counting.
Crowd counting is a particularly demanding task due to the inherent challenges, such as high-density crowds, occlusions, and scale variations. Thus, it provides a robust testbed to evaluate CLIP’s applicability to regression problems.
\begin{figure}[t]
    \centering
    \includegraphics[width=\linewidth]{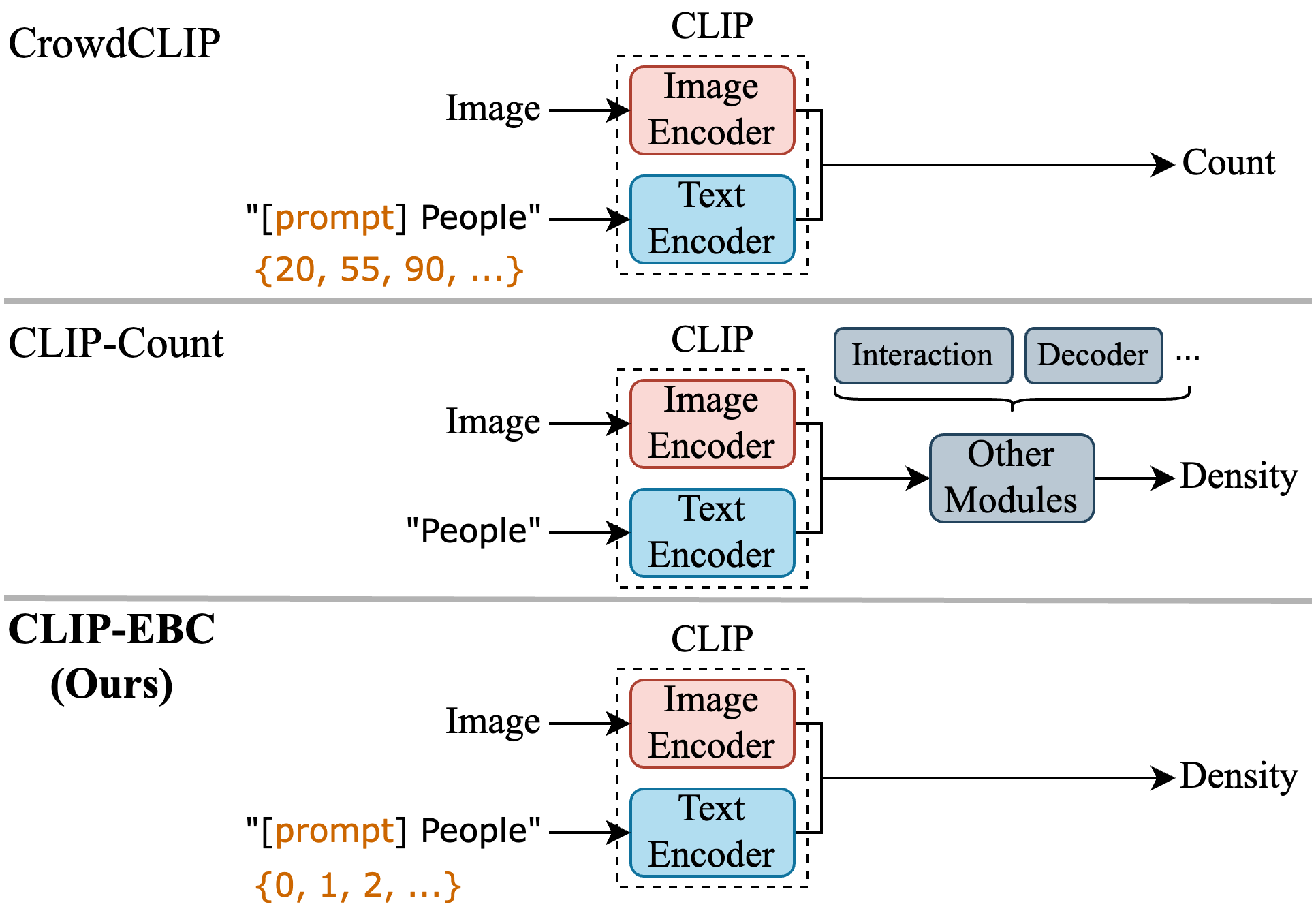}
    \caption{Comparison of our model CLIP-EBC and other CLIP-based crowd counting methods.
    \textbf{Top:} CrowdCLIP \cite{liang2023crowdclip} predicts total counts from global image features. Due to the incapability to capture spatial details, it suffers from reduced accuracy.
    \textbf{Middle:} CLIP-Count \cite{jiang2023clip} employs extra modules to generate density maps, improving accuracy at the cost of increased complexity.
    \textbf{Bottom:} CLIP-EBC is the first fully CLIP-based density-map approach, achieving high accuracy without additional modules.
    }
    \label{fig:comparison}
    \vspace{-12pt}
\end{figure}
Crowd counting is typically formulated as a blockwise density map regression task \cite{li2018csrnet, wang2020distribution, ma2022fusioncount}.
Therefore, an intuitive approach to adapt CLIP for crowd counting is through classification frameworks, where blockwise count values are quantized into discrete bins.
However, directly applying CLIP in existing classification frameworks \cite{liu2019counting, xiong2019open} presents significant challenges.

\begin{figure*}[tbp]
    \centering
    \includegraphics[width=\textwidth]{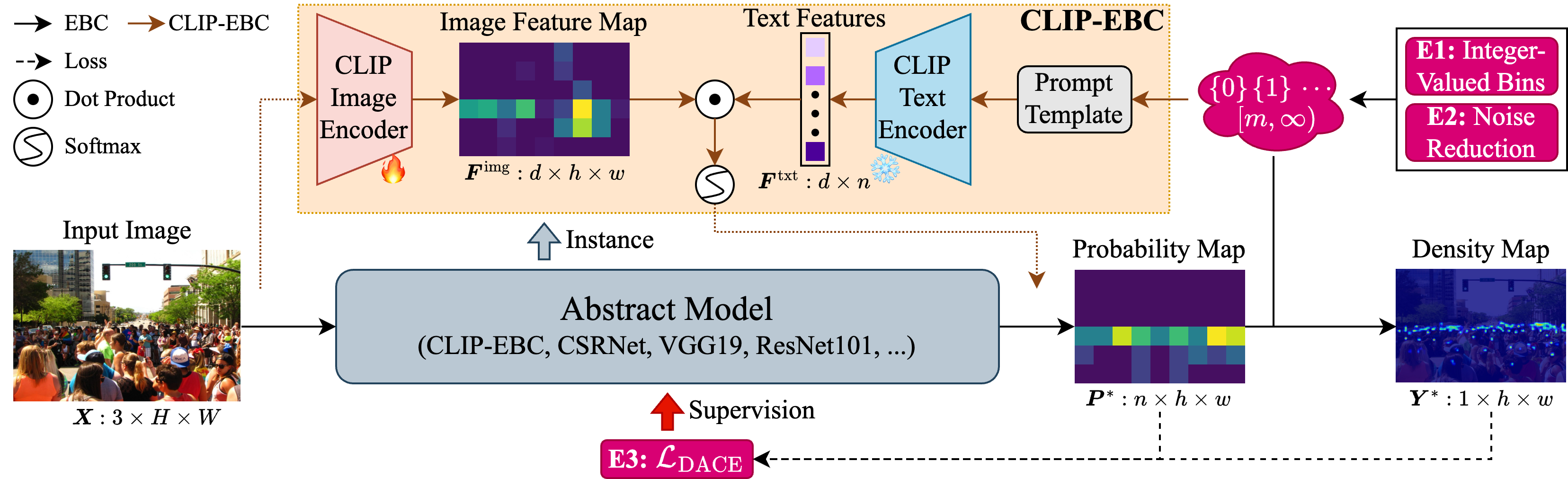}
    \caption{
    Overview of the EBC framework and the CLIP-EBC model.
    The EBC framework introduces an integer-based quantization strategy and noise reduction to partition count values into $n$ predefined bins (e.g., $\{0\}, \{1\}, \cdots, [m, \infty)$).
    Given an input image, the crowd counting model implemented under EBC generates a blockwise probability map.
    The density map is obtained by averaging the bin midpoints weighted by the probability map.
    Finally, the predicted probability map and the density map are fed into the proposed $\mathcal{L}_\text{DACE}$ loss (Eq. \eqref{eqn:loss}), which is used to fine-tune the model.
    CLIP-EBC, an instance of EBC, leverages the predefined bins and the prompt template to construct $n$ text prompts (e.g., ``There is 0 person'').
    The frozen CLIP \cite{radford2021learning} text encoder extracts text embeddings, while the fine-tuned CLIP image encoder generates image feature maps. 
    For the image feature vector at each spatial location, we calculate its cosine similarity with the text embedding for each bin and then apply softmax to obtain the probability map.
    Subsequent steps align with the EBC framework to generate the final outputs.
    }
    \label{fig:model}
    \vspace{-8pt}
\end{figure*}

For instance, existing classification methods employ Gaussian smoothing to reduce the sparsity of the ground-truth density maps, transforming the initially discrete count value spaces $\mathbb{N}$ into a continuous space $\mathbb{R}_+$.
This requires dividing the space into consecutive real-valued intervals to construct classification bins. 
While such bins can cover $\mathbb{R}_+$, they are poorly suited for constructing CLIP-compatible text prompts, since CLIP’s text encoder is pre-trained on natural language corpora, where descriptions are typically human-readable and discrete (e.g., ``there are 3 people'').
Prompts involving real values (e.g., ``there are 3.14 people'') lack semantic clarity and may result in noisy text embeddings and degraded performance. 
Besides, the use of adjacent bins introduces label ambiguity, particularly for samples near shared boundaries. For example, the bins [0, 0.05) and [0.05, 0.1) in \cite{liu2019counting} create uncertainty for a block with a count of 0.05, as it is unclear which bin it should belong to. Such ambiguity weakens the reliability of predictions.
Another limitation of current classification-based methods is their sole focus on optimizing the classification accuracy. This can lead to suboptimal performance on regression metrics, such as Mean Absolute Error (MAE) or Root Mean Squared Error (RMSE), which are critical for counting tasks.

To address these challenges, we propose the \textbf{E}nhanced \textbf{B}lockwise \textbf{C}lassification (\textbf{EBC}) framework, specifically designed to overcome the limitations of current classification-based methods.
Our EBC framework is model-agnostic: existing regression-based methods can be implemented using EBC with minimal modifications to achieve substantial performance improvements.
Building on this framework, we introduce CLIP-EBC, which fully leverages the original structure of CLIP for crowd counting.
Unlike other CLIP-based methods \cite{liang2023crowdclip, jiang2023clip} (see Fig.~\ref{fig:comparison}), CLIP-EBC is the first model to fully utilize CLIP for density-map-based crowd counting.
While CrowdCLIP \cite{liang2023crowdclip} also relies solely on CLIP, it predicts total counts instead of density maps, limiting its ability to capture fine-grained spatial details for accurate predictions.

Experiments across four benchmark datasets demonstrate the significant advantages of our EBC framework. For instance, EBC reduces Mean Absolute Error (MAE) by up to 44.5\% on the UCF-QNRF dataset \cite{idrees2018composition} compared to the standard blockwise classification method \cite{liu2019counting}. Furthermore, CLIP-EBC surpasses state-of-the-art crowd counting methods, achieving an MAE of 58.2 and a Root Mean Square Error (RMSE) of 268.5 on the NWPU-Crowd test split \cite{wang2020nwpu}. These results demonstrate that CLIP, with the support of the EBC framework, can accurately estimate crowd density distributions. To summarize, our contributions are as follows:
\begin{itemize}
    \item We propose the novel \textbf{E}nhanced \textbf{B}lockwise \textbf{C}lassification (\textbf{EBC}) framework that substantially improves over the classification-based methods by addressing issues related to Gaussian smoothing and loss functions.
    \item Building upon EBC, we present \textit{the first fully CLIP-based crowd counting model} \textbf{CLIP-EBC} that is capable of accurately estimating crowd densities.
    \item We conduct extensive experiments across multiple data-sets to showcase the effectiveness of EBC in enhancing existing methods and the competence of CLIP-EBC as a state-of-the-art crowd-counting method.
\end{itemize}
\section{Related Work}
\label{sec:related_work}
%
%
%
\noindent\textbf{Classification-Based Crowd Counting Methods.}
These models aim to address the long-tail distribution of count values, where large counts are underrepresented, hindering the performance of regression-based models.
These methods partition the range $[0, \infty)$ into consecutive non-overlapping intervals to increase the sample size per class.
For instance, Xiong \etal \cite{xiong2019open} proposed DCNet, which predicts counts at multiple levels using a shared set of bins.
However, this approach fails to account for the lower likelihood of large counts at lower levels, exacerbating class imbalance.
Liu \etal \cite{liu2019counting} addressed this by introducing blockwise classification, where models output probability maps.
However, these methods rely on Gaussian smoothing, which introduces label noise and boundary ambiguities, leading to reduced accuracy.
Also, optimizing for classification accuracy alone cannot guarantee optimal performance under regression metrics such as MAE.

%
\noindent\textbf{CLIP in Crowd Counting.}
CLIP \cite{radford2021learning} has demonstrated exceptional performance in recognition-related tasks, such as zero-shot image classification \cite{radford2021learning}, 
and semantic segmentation \cite{luddecke2022image},
but there are only few studies in regression-related tasks, particularly crowd counting.
Liang \etal \cite{liang2023crowdclip} proposed CrowdCLIP, an unsupervised (or weakly-supervised) fully CLIP-based model, which operates on global image features extracted by CLIP’s image encoder. 
This global representation lacks spatial granularity and cannot capture the fine details or density variations across different parts of the image.
Jiang \etal \cite{jiang2023clip} extended CLIP for text-guided zero-shot object counting, incorporating additional modules such as multimodal interaction layers and decoders to generate density maps.
In contrast, we propose CLIP-EBC, the first crowd counting framework fully based on CLIP, designed to estimate crowd density maps without relying on external modules. Fig.~\ref{fig:comparison} presents the comparison with these methods .
\section{Methods}
\label{sec:method}
Our \textbf{E}nhanced \textbf{B}lockwise \textbf{C}lassification (\textbf{EBC}) framework improves the Standard Blockwise Classification (SBC) \cite{liu2019counting} from three key aspects: the quantization strategy, noise control, and loss formulation.
In this section, we first briefly recapitulate SBC, and then provide a detailed explanation of our enhancements.
Finally, we demonstrate how to leverage our model-agnostic EBC framework with CLIP for density map estimation, introducing our model \textbf{CLIP-EBC}. Fig.~\ref{fig:model} illustrates the overview of EBC and CLIP-EBC.
\subsection{Enhanced Blockwise Classification}
\label{sec:ebc}
\subsubsection{Recap: Standard Blockwise Classification (SBC)}
Similar to many regression-based methods \cite{li2018csrnet, liu2019counting, wang2020distribution, ma2022fusioncount}, SBC also employs blockwise prediction.
Formally, given an input image
$\boldsymbol{X} \in \mathbb{R}_+^{3 \times H \times W}$,
with $3$ 
channels, height $H$ and width $W$, 
SBC outputs a probability map
\begin{equation}\label{eqn:prob_map}
\boldsymbol{P}^* \in \mathbb{R}_+^{(n, \, H // r, \, W // r)},
\end{equation}
where $//$ represents the floor division operator and the integer $r$ is a model-related 
reduction factor.
At each spatial location $(i, \, j)$, the $n$-dimensional vector $\boldsymbol{P}^*_{:, \, i, \, j}$ represents the predicted probability distribution of the count value in the block $(r(i-1): ri, \, r(j-1): rj)$ of the image over the $n$ predefined bins $\{\mathcal{B}_i \mid i = 1, \cdots, n\}$.
%
These bins satisfy 
\begin{equation*}
    \forall p \neq q, \, \mathcal{B}_p \cap \mathcal{B}_q = \emptyset \quad\text{and}\quad \mathcal{S} \subseteq \cup_{i=1}^{n} \mathcal{B}_i,
\end{equation*}
where $\mathcal{S}$ denotes the set of all possible count values.
The predicted density map $\boldsymbol{Y}^*$ can be generated by taking the expectation or the mode over the bins. In the former case, $\boldsymbol{Y}^*$ is calculated by
\begin{equation}\label{eqn:expectation}
    \boldsymbol{Y}^*_{i,j} = \sum_{k=1}^n \boldsymbol{P}^*_{k,i,j} \cdot b_k,
\end{equation}
where $b_k$ is the average count value in the bin $\mathcal{B}_k$.

In SBC, Gaussian smoothing is used for preprocessing the ground truth density maps to reduce sparsity, transforming $\mathcal{S}$ from an initially discrete space $\mathbb{N}$ into a continuous space $\mathbb{R}_+$.
Consequently, a sequence of bordering real-valued intervals is employed as bins to cover $\mathcal{S} \subseteq \mathbb{R}_+$.
This quantization strategy leads to inconsistency with CLIP's pre-training corpus and makes samples with count values near the interval borders difficult to classify.
For instance, it is challenging for models to determine whether to classify a block with a count value of 0.05 into $\mathcal{B}_1 = [0, \, 0.05)$ or $\mathcal{B}_2 = [0.05, \, 0.1)$.
Moreover, because the spatial size of each person is not provided, Gaussian smoothing introduces noise into the ground-truth density maps, negatively impacting the model's generalizability \cite{wang2020distribution}.
\subsubsection{Enhancement 1: Integer-Valued Bins}
To avoid the issues caused by Gaussian smoothing, we propose to bypass this technique in preprocessing and preserve the discreteness of the original count space $\mathcal{S} \subseteq \mathbb{N}$.
Correspondingly, we introduce three integer-valued binning strategies: \textit{fine}, \textit{dynamic}, and \textit{coarse}.
At the fine level, each bin $\mathcal{B}_k$ contains only one integer, which also serves as the average count value $b_k$.
Thus, this policy provides bins with unbiased average count values.
For coarse-level bins, each bin comprises two integers to increase the sample size, excluding 0, which forms a bin by itself. This approach mitigates the long-tail distribution problem but also introduces biases in the average count values. Let us take the bin $\{1, \, 2\}$ and its average count value 1.5 as an example: when the ground-truth count value is 1 and the predicted probability score over this bin is $1.0$, the predicted count value is thus $1.5 \times 1.0 = 1.5$, a bias of 0.5 from the ground truth.
The dynamic binning policy is a middle ground between the fine and coarse policies. It assigns small count values to individual bins, while every two large count values are merged into a single bin. This strategy balances between lowering the biases and increasing the sample sizes.
\begin{figure}[t]
\begin{subfigure}{0.49\linewidth}
    \includegraphics[width=\textwidth]{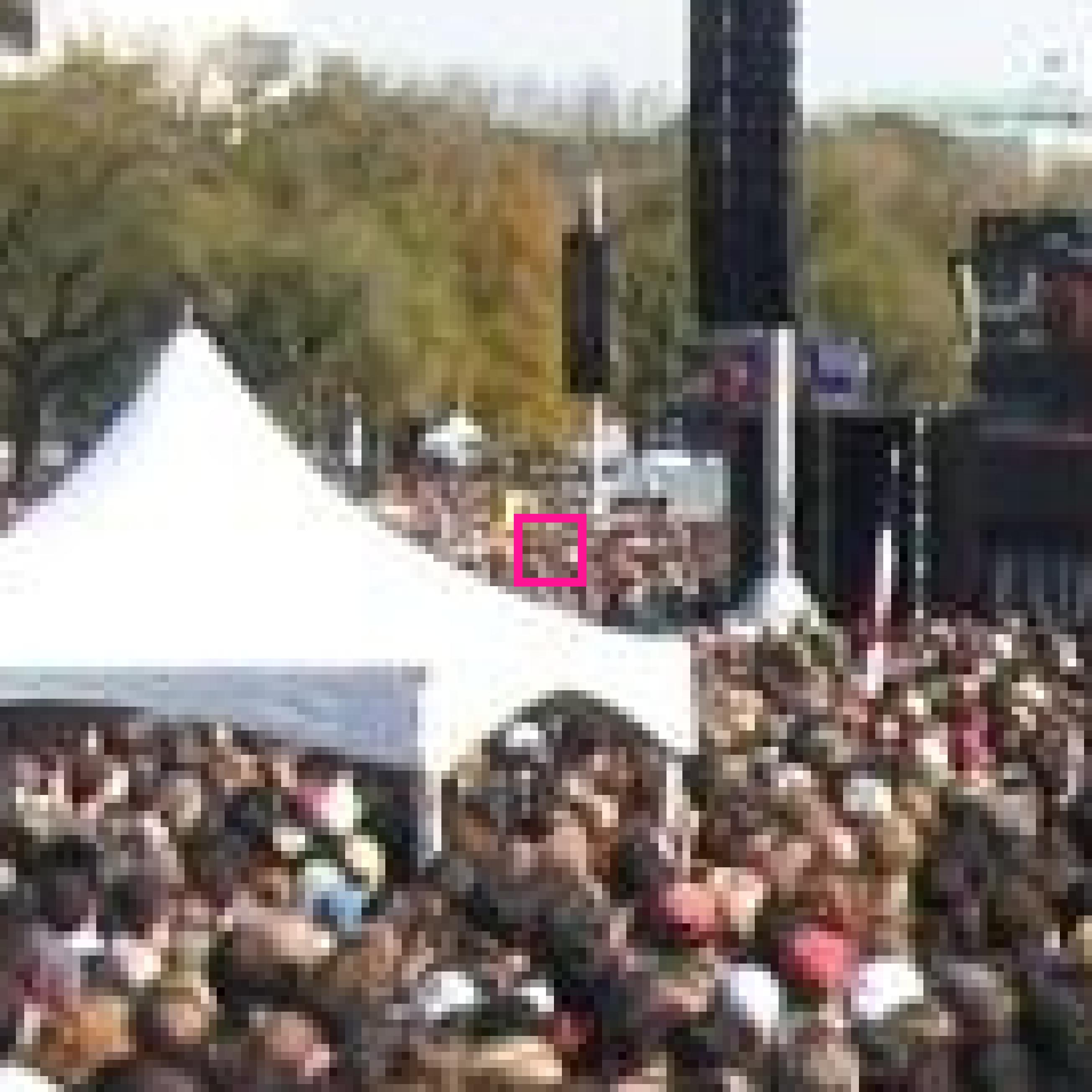}
    \caption{Region with noisy labels}
    \label{fig:dense_2}
\end{subfigure}
\hfill
\begin{subfigure}{0.49\linewidth}
    \includegraphics[width=\textwidth]{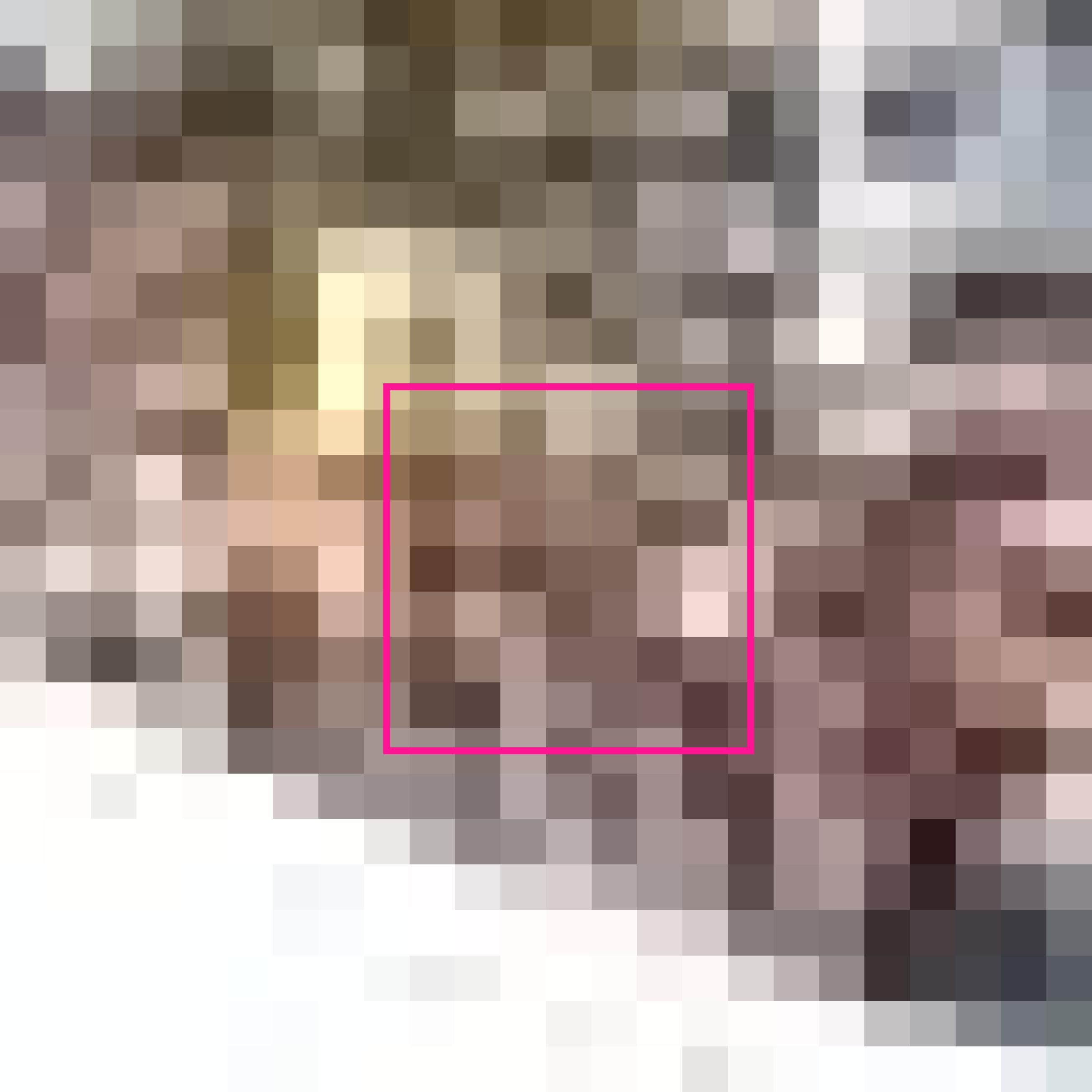}
    \caption{Zoomed-in view of the region}
    \label{fig:dense_3}
\end{subfigure}
\caption{
An example from ShanghaiTech A dataset \cite{zhang2016single} illustrating an extremely dense area with erroneous annotations.
The $8 \times 8$ magenta box in Fig.~\ref{fig:dense_2} highlights a congested region with an labeled count of \textcolor{red}{9} individuals. However, the zoomed-in view in Fig. \ref{fig:dense_3} reveals that no discernible human figures can be identified within the marked box, evidencing label noise in highly dense regions.
%
}
\label{fig:dense}
\vspace{-10pt}
\end{figure}
\begin{table}[t]
\centering
\setlength{\tabcolsep}{1mm}
\begin{tabular}{ccc|c}
\toprule
\textbf{\begin{tabular}[c]{@{}c@{}}Integer-Valued Bins\\(E1)\end{tabular}} & \textbf{\begin{tabular}[c]{@{}c@{}}Noise Reduction\\(E2)\end{tabular}} & \textbf{\begin{tabular}[c]{@{}c@{}}Combined Loss\\(E3)\end{tabular}} &
  \textbf{MAE} $\downarrow$ \\ \midrule
\xmark & \xmark & \xmark         & 140.6 \\
\cmark & \xmark & \xmark         & 88.3  \\
\cmark & \cmark & \xmark         & 85.8  \\ \midrule
\cmark & \cmark & \cmark ($\lambda=0.01$) & 83.9  \\
\cmark & \cmark & \cmark ($\lambda=0.10$) & 83.4  \\
\cmark & \cmark & \cmark ($\lambda=1.00$) & \textbf{77.9} \\
\cmark & \cmark & \cmark ($\lambda=2.00$) & 82.3  \\ 
\bottomrule
\end{tabular}
\caption{
Influence of each enhancement in our \textbf{EBC} framework evaluated on UCF-QNRF \cite{idrees2018composition}.
When no enhancement is utilized, the model deteriorates into SBC \cite{liu2019counting}.
The reduced mean absolute errors demonstrate the effectiveness of our enhancement strategies and the superiority of EBC.
}
\label{tab:ebc_ablation}
\vspace{-10pt}
\end{table}
\subsubsection{Enhancement 2: Noise Reduction}
Another challenge often overlooked by existing approaches is the presence of erroneous and noisy annotations within highly dense crowd areas, as illustrated in Fig.~\ref{fig:dense}.
%
This issue can stem from two primary factors: 
1) low-resolution images, where annotators struggle to precisely determine the head counts in congested areas;
and 2) datasets with images resized to significantly smaller resolutions after labeling to optimize storage and training time.
For instance, consider an HD image annotated with 1000 people, but subsequently resized to $64 \times 64$ pixels.
In such a scenario, each individual occupies an average of only 4 pixels––insufficient to distinctly represent a human figure.
To address this problem, we propose a noise reduction strategy by merging all count values equal to or greater than a preset threshold $m$ into a single large bin. 
This approach is integrated into our three binning strategies as follows:
\begin{align}
    \textbf{Fine:} & \, \{0\}, \{1\}, \cdots, [m, \infty) \label{eqn:fine_bins} \\
    \textbf{Coarse:} & \, \{0\}, \{1, 2\}, \cdots, [m, \infty) \label{eqn:coarse_bins} \\
    \textbf{Dynamic:} & \, \{0\}, \cdots, \{n\}, \{n+1, n+2\}, \cdots, [m, \infty) \label{eqn:dynamic bins}
\end{align}
An additional advantage of this strategy is that it increases the sample size of large count values by merging them into one bin, thereby alleviating the long-tail distribution issue.
\subsubsection{Enhancement 3: Combined Loss.}
Previous blockwise classification methods \cite{liu2019counting, xiong2019open} primarily focus on minimizing the cross-entropy loss between the predicted probability map $\boldsymbol{P}^*$ and the ground truth $\boldsymbol{P}$ but often overlook the mismatch in count values.
Since two different probability distributions can yield the same classification error while having different expectations (or modes) for the predicted count values, models trained with such loss functions may not guarantee accurate count estimates.
To address this limitation, we propose the \textbf{D}istance-\textbf{A}ware-\textbf{C}ross-\textbf{E}ntropy (\textbf{DACE}) loss:
\begin{align}
    \label{eqn:loss}
    \mathcal{L}_\text{DACE} = & \mathcal{L}_\text{class} (\boldsymbol{P}^*, \, \boldsymbol{P}) + \lambda \mathcal{L}_\text{count} (\boldsymbol{Y}^*, \, \boldsymbol{Y}) \notag \\
    = & - \sum_{i=1}^{H // r} \sum_{j = 1}^{W // r} \sum_{k = 1}^n \boldsymbol{1}(\boldsymbol{P}_{k, i, j} = 1) \log \boldsymbol{P}^*_{k, i, j} \notag \\ &+ \lambda \mathcal{L}_\text{count} (\boldsymbol{Y}^*, \, \boldsymbol{Y}),
\end{align}
where $\boldsymbol{1}$ is the indicator function, $\boldsymbol{P}$ is the one-hot encoded ground-truth probability map, $\boldsymbol{P}^*$ is the predicted probability map, $\boldsymbol{Y}$ is the ground-truth density map, $\boldsymbol{Y}^*$ is the predicted density map, and $\lambda$ is a weighting factor balancing $\mathcal{L}_\text{class}$ and $\mathcal{L}_\text{count}$.
Since the ground-truth density maps are not smoothed, we use the DMCount \cite{wang2020distribution} based on the discrete Wasserstein distance as the count loss $\mathcal{L}_\text{count}$.
\begin{table}[t]
\centering
\small{
\begin{tabular}{l|c|ccc}
\toprule
\textbf{Backbone} &
  \textbf{Type} &
  \textbf{\begin{tabular}[c]{@{}c@{}}MAE\\ (Orig.)\end{tabular}} &
  \textbf{\begin{tabular}[c]{@{}c@{}}MAE\\ (EBC)\end{tabular}} &
  \textbf{Improvement} \\ \midrule
VGG16              & C    & 140.6       & 77.9      & \textbf{44.5\%}      \\
ResNet101          & C    & 127.3       & 79.7      & 37.3\%      \\
MobileNetV2        & C    & 135.9       & 83.2      & 38.7\%      \\
DenseNet201        & C    & 118.1       & 82.9      & 29.8\%      \\ \midrule
CSRNet            & R    & 119.2       & 79.3      & 33.4\%      \\
DMCount         & R    & \textbf{85.6}        & \textbf{77.2}      & 9.8\%       \\ \bottomrule
\end{tabular}
}
\caption{
Our EBC framework can improve both classifica-tion-based and regression-based methods. We reimplement both SBC models \cite{liu2019counting} and two regression-based methods (CSRNet \cite{li2018csrnet} and DMCount \cite{wang2020distribution}) using our EBC framework on the UCF-QNRF dataset. Regardless of the choice of the backbone, our EBC always improves the performance.
}
\label{tab:backbones}
\vspace{-10pt}
\end{table}
\subsection{The Structure of CLIP-EBC}
As CLIP-EBC is implemented under our proposed EBC framework, here we focus on how to generate the predicted probability map $\boldsymbol{P}^*$. Additional essential details, such as the loss function used in training, are provided in Section \ref{sec:ebc}.

The image encoder of CLIP-EBC comprises a feature extractor $\mathcal{F}$, an interpolation layer $\mathcal{U}$, and a projection layer $\mathcal{P}$. 
We utilize CLIP's image encoder without the final pooling and projection operations to extract the feature map
\begin{equation}
    \boldsymbol{H} = \mathcal{F} (\boldsymbol{X}) \in \mathbb{R}^{c \times (H // p) \times {W // p}},
\end{equation}
where $c$ represents the number of output channels (for the ResNet backbone) or the embedding dimension (for the ViT backbone), and $p$ represents the model reduction factor (ResNet) or the patch size (ViT).
Subsequently, $\boldsymbol{H}$ is passed to $\mathcal{U}$ to adjust the spatial size
\begin{equation}
    \tilde{\boldsymbol{H}} = \mathcal{U} (\boldsymbol{H}) \in \mathbb{R}^{c \times (H // r) \times {W // r}}.
\end{equation}
Finally, we employ one $1 \times 1$ convolutional layer $\mathcal{P}$ to project $\tilde{\boldsymbol{H}}$ into the CLIP embedding space, yielding
\begin{equation}
    \boldsymbol{F}^\text{img} = \mathcal{P} (\tilde{\boldsymbol{H}}) \in \mathbb{R}^{d \times (H // r) \times {W // r}},
\end{equation}
where $d$ is the dimension of the CLIP embedding space.
\begin{table*}[t]
\centering
\begin{tabular}{l|c|cccccccc}
\toprule
\multirow{2}{*}{\textbf{Methods}} &
\multirow{2}{*}{\textbf{Type}} &
\multicolumn{2}{c}{\textbf{ShanghaiTech A}} &
\multicolumn{2}{c}{\textbf{ShanghaiTech B}} &
\multicolumn{2}{c}{\textbf{UCF-QNRF}} &
\multicolumn{2}{c}{\textbf{NWPU (Val)}} \\
&
&
\textbf{MAE} &
\textbf{RMSE} &
\textbf{MAE} &
\textbf{RMSE} &
\textbf{MAE} &
\textbf{RMSE} &
\textbf{MAE} &
\textbf{RMSE} \\ \midrule
  CrowdCLIP \cite{liang2023crowdclip} &
  U &
  146.1 &
  236.3 &
  69.3 &
  85.8 &
  283.3 &
  488.7 &
  - &
  - \\ \midrule\midrule
  CLIP-Count\textsuperscript{\dag} \cite{jiang2023clip} &
  R &
  192.6 &
  308.4 &
  45.7 &
  77.4 &
  - &
  - &
  - &
  - \\
  DMCount \cite{wang2020distribution} &
  R &
  59.7 &
  95.7 &
  7.4 &
  11.8 &
  85.6 &
  148.3 &
  70.5 &
  357.6 \\
  P2PNet \cite{song2021rethinking} &
  R &
  \underline{52.7} &
  \underline{85.0} &
  {6.3} &
  \underline{9.9} &
  85.3 &
  154.5 &
  77.4 &
  362.0 \\
  CLTR \cite{liang2022end} &
  R &
  56.9 &
  95.2 &
  6.5 &
  10.2 &
  85.8 &
  141.3 &
  61.9 &
  246.3 \\
  STEERER \cite{han2023steerer} &
  R &
  54.5 &
  86.9 &
  \textbf{5.8} &
  \textbf{8.5} &
  \textbf{74.3} &
  \textbf{128.3} &
  - &
  - \\ \midrule \midrule
  S-DCNet \cite{xiong2019open} &
  C &
  58.3 &
  95.0 &
  6.7 &
  10.7 &
  104.4 &
  176.1 &
  - &
  - \\
  UEPNet \cite{wang2021uniformity} &
  C &
  54.6 &
  91.2 &
  6.4 &
  10.9 &
  81.1 &
  \underline{131.6} &
  - &
  - \\
  \textbf{CLIP-EBC (ResNet50, ours)} &
  C &
  54.0 &
  \textbf{83.2} &
  \underline{6.0} &
  10.1 &
  80.5 &
  136.6 &
  \underline{38.6} &
  \underline{90.3} \\
  \textbf{CLIP-EBC (ViT/B-16, ours)} &
  C &
  \textbf{52.5} &
  {85.9} &
  6.6 &
  10.5 &
  \underline{80.3} &
  139.3 &
  \textbf{36.6} &
  \textbf{81.7} \\ \bottomrule
\end{tabular}
\caption{
Comparison of our model \textbf{CLIP-EBC} with recent crowd counting approaches on ShanghaiTech A \& B \cite{zhang2016single}, UCF-QNRF \cite{idrees2018composition} and NWPU-Crowd (Val) \cite{wang2020nwpu}. The ``Type'' column groups all methods into unsupervised (``U''), regression-based (``R''), or classification-based (``C''). The best results are highlighted in \textbf{bold}, and the second best results are \underline{underlined}. Our model CLIP-EBC demonstrates competitive performance across all datasets, with particularly significant superiority on the NWPU validation split, where it outperforms all existing methods. \textsuperscript{\dag}: Results of cross-dataset evaluation.
}
\label{tab:sota}
\vspace{-8pt}
\end{table*}
For text feature extraction, we start by generating the input text prompts. Given a set of bins $\{\mathcal{B}_i \mid i = 1, \cdots, n \}$, we create a text prompt for each bin $\mathcal{B}_i$ according to the following rules:
\begin{itemize}
    \item If $\mathcal{B}_i = \{b_i\}$, the text prompt is ``\texttt{There is/are $b_i$ person/people}'', with the choice between ``is/are'' and ``person/people'' determined by whether $b_i > 1$.
    \item If $\mathcal{B}_i$ is finite and contains more than one element, then the text prompt becomes ``\texttt{There is/are between $\min (\mathcal{B}_i)$ and $\max (\mathcal{B}_i)$ person/people}''. The decision between ``is/are'' and ``person/people'' is again made to ensure grammatical correctness.
    \item If $\mathcal{B}_i = [m, \infty)$, then the text prompt is ``\texttt{There are more than $m$ people}''.
\end{itemize}
Next, the resulting $n$ text prompts are tokenized using CLIP's tokenizer.
We then input the tokenized text into the original CLIP text encoder, with its parameters frozen during training. This process yields text features $\boldsymbol{F}^\text{txt}$ with dimensions $d \times n$.
The probability map $\boldsymbol{P}^*$ can be computed from the image feature map $\boldsymbol{F}^\text{img}$ and the text features $\boldsymbol{F}^\text{txt}$. First, we calculate the cosine similarity between the image feature vector $\boldsymbol{F}^\text{img}_{:, \, i,\, j}$ and the $n$ extracted text embeddings. We then normalize these similarities using softmax to obtain $\boldsymbol{P}^*_{:, \, i, \, j}$, which denotes the probability scores across the $n$ bins for block $(i, \, j)$. To obtain the predicted density map $\boldsymbol{Y}^*$, we use Eq.~\eqref{eqn:expectation}. We train CLIP-EBC with our loss function in Eq.~\eqref{eqn:loss}.
\section{Experiments}
\label{sec:exp}
\textbf{Implementation Details.} We set the block size $r = 8$ (unless explicitly specified), which is commonly used in many works \cite{li2018csrnet, liu2019counting, wang2020distribution}. For this block size, we set $m = 4$. We use the Adam optimizer \cite{kingma2014adam} to train all our models with an initial learning rate of $1e-4$, which is adjusted through a cosine annealing schedule. The batch size is fixed at $8$ for all datasets. For our ViT-based CLIP-EBC model, we apply visual prompt tuning \cite{jia2022visual} by prepending 32 learnable tokens to each layer, rather than fine-tuning the entire backbone. Following existing methods, we use the mean absolute error (MAE) and root mean square error (RMSE) to evaluate our models.
%
\subsection{Effectiveness of EBC}
We first demonstrate the effectiveness of our EBC framework using VGG16 \cite{simonyan2015very} as the backbone, trained and evaluated on the UCF-QNRF dataset.
The results are shown in Table~\ref{tab:ebc_ablation}. The SBC framework (i.e., the scenario where no enhancement is used) achieves a mean absolute error (MAE) of 140.6.
By utilizing integer-valued bins (\textbf{E1}), the MAE is remarkably reduced from 140.6 to 88.3--a significant 37.1\% improvement.
This outcome underscores the effectiveness of our approach in reducing ambiguity near the borders.
By additionally adopting noise reduction (\textbf{E2}), the MAE further decreases to 85.8.
Performance improves even more when the combined loss (\textbf{E3}) is also used.
Specifically, increasing the value of $\lambda$ from 0.00 to 1.00 results in improved performance, with the optimal result (77.9) achieved with $\lambda = 1.00$.

We also verify that the performance improvement of EBC is agnostic to the choice of backbones.
We re-implement three other backbones (ResNet101 \cite{he2016deep}, MobileNetV2 \cite{sandler2018mobilenetv2} and DenseNet201 \cite{huang2017densely}) used by the SBC paper \cite{liu2019counting}, as well as two classic regression-based methods (CSRNet \cite{li2018csrnet} and DMCount \cite{wang2020distribution}) within our EBC framework.
The results in Table~\ref{tab:backbones} show that our EBC framework consistently yields improved performance (up to 44.5\%) across different backbones.
\subsection{CLIP-EBC Compared with State-of-the-Art}
\begin{figure}[t]
    \centering
    \includegraphics[width=\linewidth]{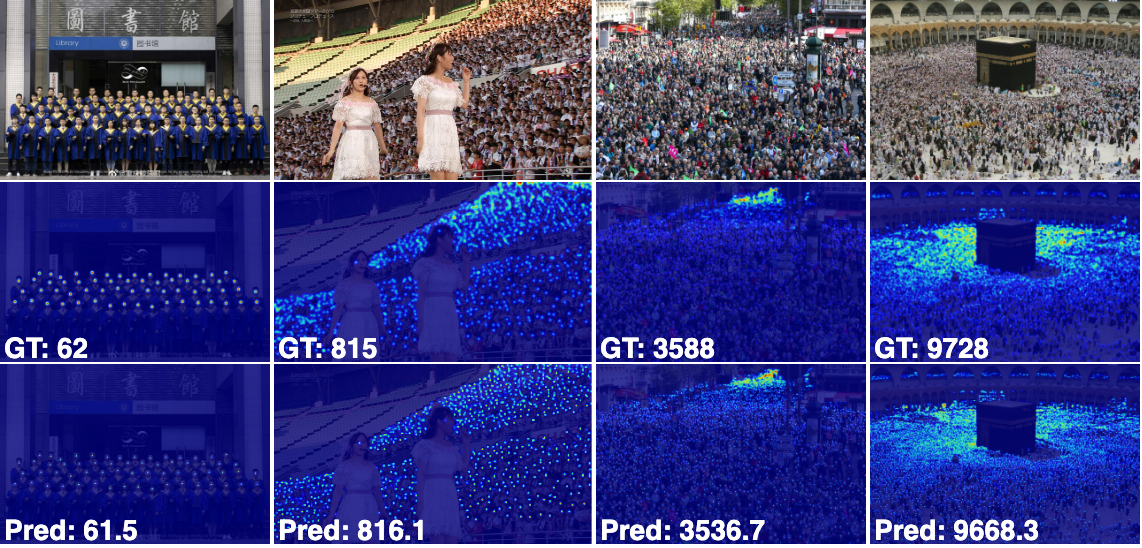}
    \caption{Visualization of density maps predicted by CLIP-EBC.}
    \label{fig:vis}
    \vspace{-10pt}
\end{figure}

\begin{table}[t]
\centering
\begin{tabular}{l|cc}
\toprule
\textbf{Methods}                   & \textbf{MAE}     & \textbf{RMSE}  \\ \midrule
P2PNet \cite{song2021rethinking}   & 72.6             & 331.6          \\
CLTR \cite{liang2022end}           & 74.4             & 333.8          \\
STEERER \cite{han2023steerer}      & {63.7} & 309.8          \\
CrowdHat \cite{wu2023boosting}     & 68.7             & {296.9}    \\
PET \cite{liu2023point}            & 74.4             & 328.5          \\ \midrule
\textbf{CLIP-EBC (ViT-B/16, ours)} & \underline{61.3}    & \underline{278.4} \\ 
\textbf{CLIP-EBC (ViT-L/14, ours)} & \textbf{58.2}    & \textbf{268.5} \\ 
\bottomrule
\end{tabular}%
\caption{Comparison of our model CLIP-EBC with state-of-the-art methods on NWPU-Crowd (Test) \cite{wang2020nwpu}. CLIP-EBC achieves both the lowest MAE and RMSE on this large-scale benchmark database, surpassing all strong baselines.}
\label{tab:nwpu_test}
\vspace{-10pt}
\end{table}
We compare CLIP-EBC with the latest crowd-counting methods.
Table~\ref{tab:sota} presents the results on the ShanghaiTech A \& B, UCF-QNRF, and NWPU-Crowd (Val) datasets, while Table~ \ref{tab:nwpu_test} illustrates the results on the test split of the NWPU-Crowd database.
Our CLIP-EBC models achieve comparable results with state-of-the-art approaches on the ShanghaiTech A \& B and UCF-QNRF datasets.
On the much larger benchmark database NWPU-Crowd, our CLIP-EBC model with the ViT-B/16 backbone achieves significant performance gains, surpassing all existing models with breakthrough results: an MAE of 36.6 and an RMSE of 81.7 on the validation split. On the test set, the ViT-B/16 backbone achieves an MAE of 61.3 and an RMSE of 278.4, outperforming all strong baselines. Moreover, the ViT-L/14 variant further reduces the errors, achieving an MAE of 58.2 and an RMSE of 268.5, setting new state-of-the-art performance. Fig.~\ref{fig:vis} provides visualized results on the validation split using CLIP-EBC with the ViT-L/14 backbone.

\section{Conclusion}
In this paper, we bridged the gap between CLIP and crowd counting by reformulating the counting task as a blockwise classification problem.
To this end, we proposed Enhanced Blockwise Classification (EBC), a novel, backbone-agnostic framework that improves upon standard blockwise classification through three key advancements: quantization, noise reduction, and loss formulation.
Within our EBC framework, we further proposed CLIP-EBC, the first fully CLIP-based crowd counting method, which successfully leverages CLIP for accurate crowd density estimation.
CLIP-EBC classifies local count values into predefined integer-valued bins by comparing the similarity between local image features and text prompt features.
%
Extensive experiments validated the effectiveness of our EBC framework, and the comparison with existing methods on four benchmark datasets demonstrated the superior performance of our CLIP-EBC model.

\bibliographystyle{IEEEbib}
\bibliography{ref}

\clearpage\appendix
\begin{figure}[htbp]
    \centering
    \includegraphics[width=\linewidth]{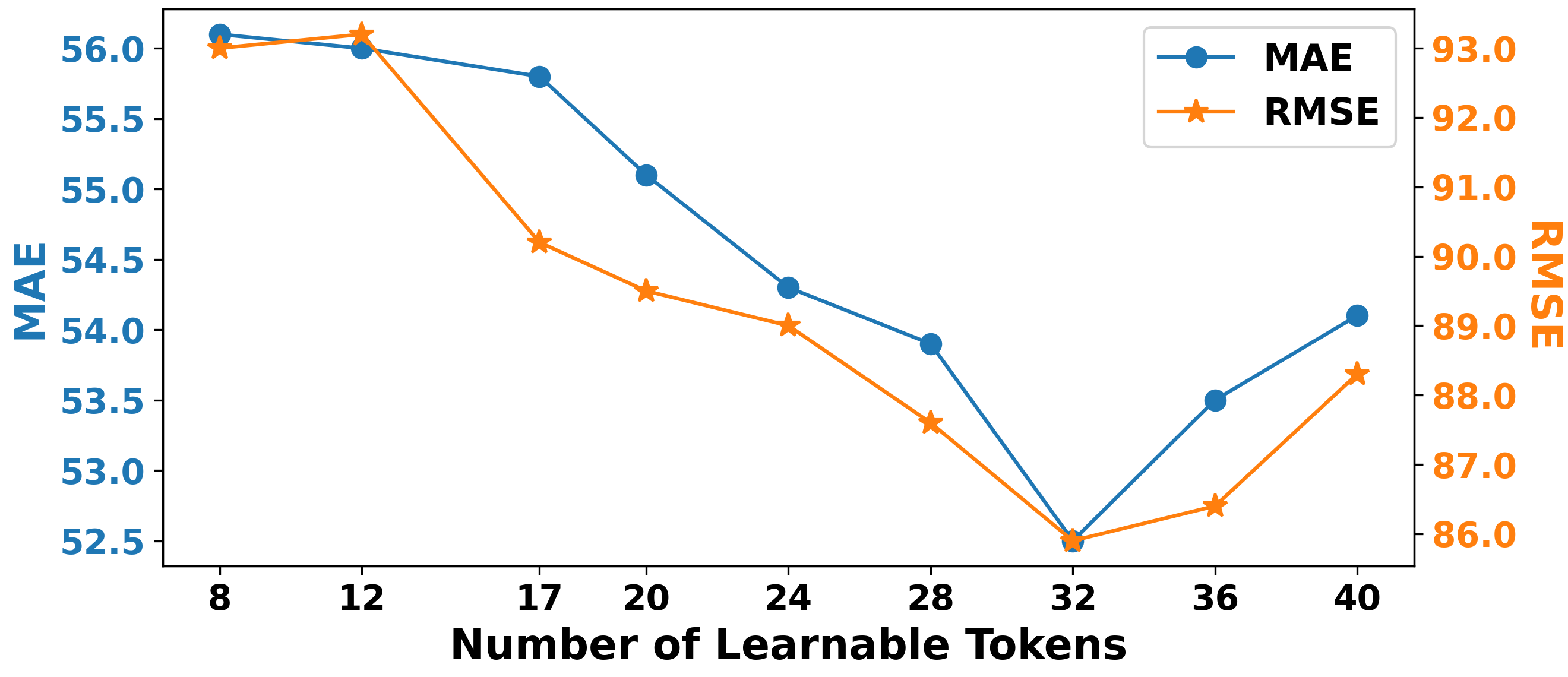}
    \caption{Influence of the number of learnable tokens in our CLIP-EBC model with the ViT-B/16 backbone. The Experiments are conducted on the ShanghaiTech A dataset. The results show that setting the number of learnable tokens to 32 yields optimal performance, achieving the lowest values for both MAE and RMSE.}
    \label{fig:vpts}
\end{figure}
\section*{Impact of the Number of Learnable Tokens}
We analyze the effect of varying the number of learnable tokens in our CLIP-EBC model with the ViT-B/16 backbone. 
The number of tokens is varied from 8 to 40, in increments of 4.
The experimental results on the ShanghaiTech A dataset are shown in Fig.~\ref{fig:vpts}.
As observed, increasing the number of learnable tokens from 8 to 32 results in a general improvement in performance, measured by both MAE and RMSE. The optimal performance is achieved at 32 tokens. 
However, beyond 32 tokens, the model's performance begins to degrade, likely due to overfitting caused by the increased learnability.
\section*{Influence of Bin Granularity}
\label{sec:exp_bin}
Given that larger blocks can accommodate a wider range of possible count values, we investigate the effect of bin granularity on the performance of the EBC framework for block sizes $r = 16$ and $r = 32$.
%
Table \ref{tab:granularity} shows the results on the ShanghaiTech A dataset using our CLIP-EBC (ViT-B/16) model.
For both block sizes, the dynamic bin policy yields the best performance, benefiting from its ability to balance between reducing biases in average count values and increasing the sample sizes for undersampled bins.
%
%
Additionally, in terms of MAE, the smaller block size $r=16$ produces better results, likely because it allows for better utilization of the spatial information present in the labels.
\section*{Data Augmentation}
\label{sec:data_aug}
We provide details of the data augmentation applied during the training in all our experiments. For each input image, we crop a region of size $su \times su$, where $s = 224$ for ViT-based models, $s = 448$ for all other models, and $u \sim \text{Uniform}[1, \, 2]$.
We then resize the cropped region to $s \times s$. This augmentation technique is designed to construct dense regions and incease the sample sizes of large bins, thereby alleviating undersampling issue caused by the long-tailed distribution of count values.
Next, we apply a random horizontal flip with a probability of 0.5, followed by random color jittering with \texttt{brightness=0.1}, \texttt{saturation=0.1}, and \texttt{hue=0.0}. The hue factor is set to 0 to avoid NaN gradient issues during training, which were observed when positive values were used.
Additionally, Gaussian blurring with a kernel size of 5 is applied to further enhance the robustness of the model
Finally, we normalize the cropped region using the ImageNet mean and standard deviation.
\begin{table}[tbp]
\centering
\begin{tabular}{l|cc|cc|cc}
\hline
\multirow{2}{*}{\textbf{Block Size}} & \multicolumn{2}{c|}{\textbf{Fine}} & \multicolumn{2}{c|}{\textbf{Dynamic}} & \multicolumn{2}{c}{\textbf{Coarse}} \\ \cline{2-7} 
         & \textbf{MAE} & \textbf{RMSE} & \textbf{MAE} & \textbf{RMSE} & \textbf{MAE} & \textbf{RMSE} \\ \hline
$r = 16$ & 76.31 & 131.74 & 75.90 & 130.48 & 77.41 & 130.48  \\
$r = 32$ & 76.94 & 130.87 & 76.06 & 127.72 & 79.39 &  132.51 \\ \hline
\end{tabular}
\caption{Effect of block sizes and bin granularity. Experiments are conducted on UCF-QNRF \cite{idrees2018composition}.
Results indicate that the dynamic bin strategy outperforms the other due, benefited from its balance between reducing biases in representation count values and increasing class sample sizes.
}
\label{tab:granularity}
\end{table}
\section*{Visualization}
\begin{figure*}[t]
\centering
\begin{subfigure}[t]{\linewidth}
    \centering
    \includegraphics[width=\textwidth]{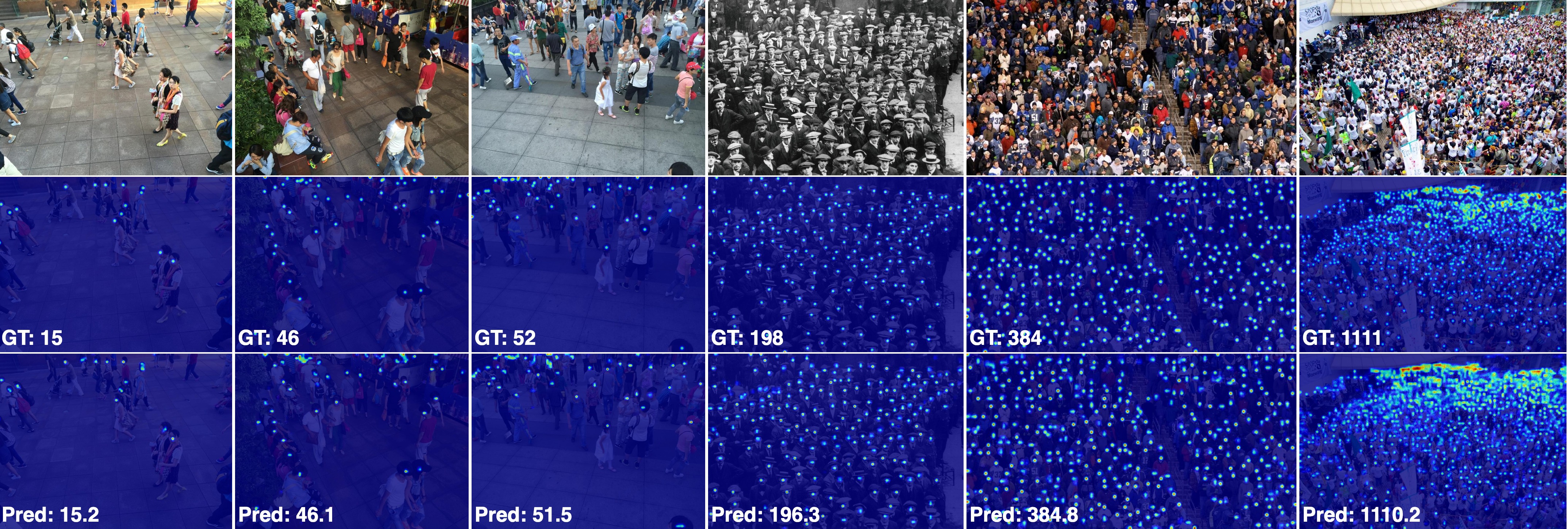}
    \caption{Visualization on the ShanghaiTech \cite{zhang2016single} dataset with our CLIP-EBC (ResNet-50) model.}
    \label{fig:vis_clip_r50}
\end{subfigure}
\begin{subfigure}[t]{\linewidth}
    \centering
    \includegraphics[width=\textwidth]{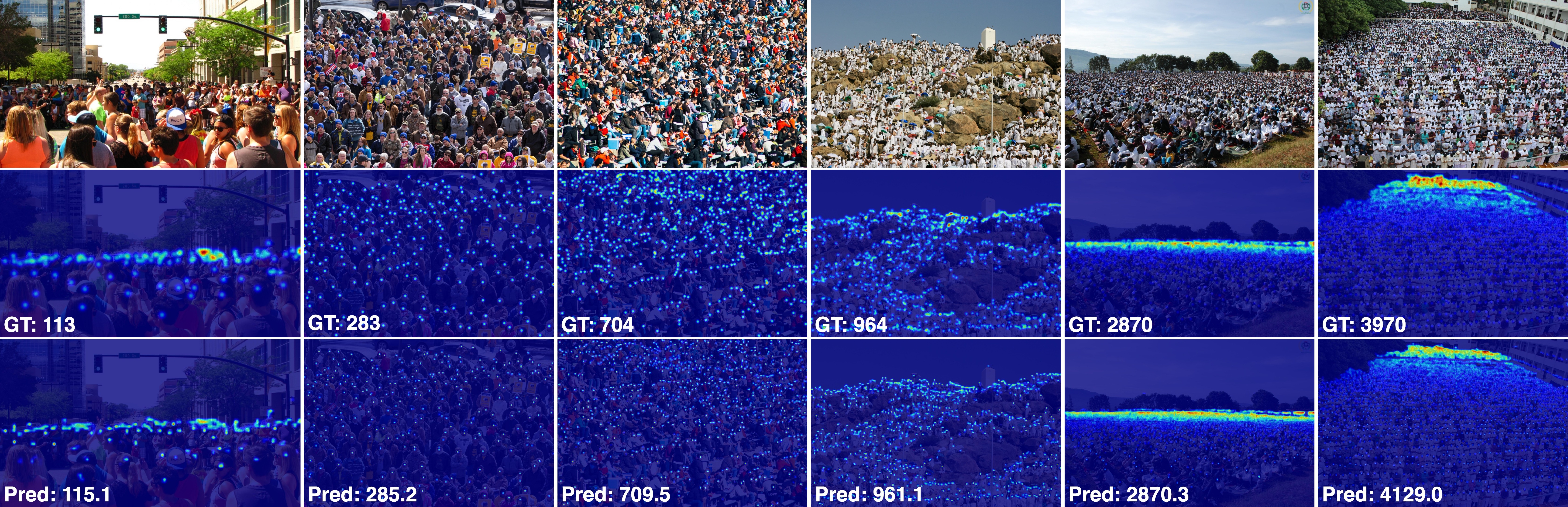}
    \caption{Visualization on the UCF-QNRF \cite{idrees2018composition} dataset with our CLIP-EBC (ViT-B/16) model.}
    \label{fig:vis_clip_vit_b}
\end{subfigure}
\begin{subfigure}[t]{\linewidth}
    \centering
    \includegraphics[width=\textwidth]{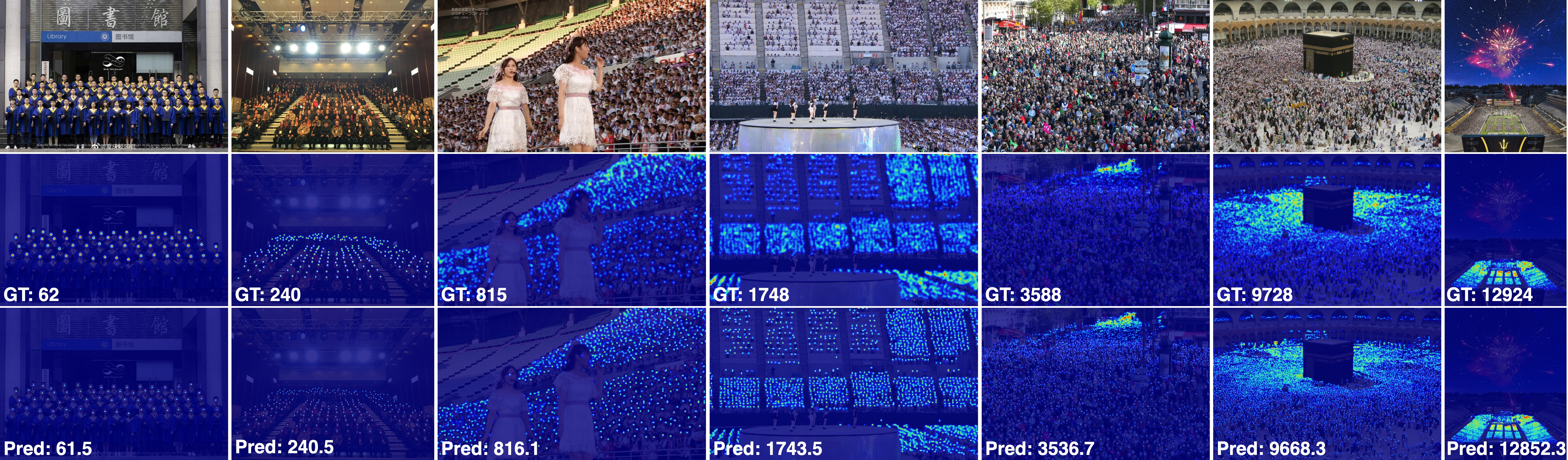}
    \caption{Visualization on the NWPU-Crowd \cite{wang2020nwpu} dataset with our CLIP-EBC (ViT-L/14) model.}
    \label{fig:vis_clip_vit_l}
\end{subfigure}
\caption{Visualization on the validation splits of different datasets using different CLIP-EBC models. Top to bottom rows: original images, the Gaussian-smoothed ground-truth density maps, and the density maps predicted by CLIP-EBC. Columns from the left to right represent scenarios of increasing density.}
\label{fig:vis_appendix}
\end{figure*}

We use our trained CLIP-EBC models to generate predicted density maps for four datasets: ShanghaiTech A \& B, UCF-QNRF, and NWPU-Crowd.
Specifically, CLIP-EBC (ResNet50) is applied to ShanghaiTech, CLIP-EBC (ViT-B/16) to UCF-QNRF, and CLIP-EBC (ViT-L/14) to NWPU-Crowd.
A curated set of 19 representative images--six from ShanghaiTech, six from UCF-QNRF, and seven from NWPU-Crowd--are selected for visualization alongside their corresponding ground-truth and predicted density maps in Fig.~\ref{fig:vis_appendix}.
\begin{itemize}
    \item Fig.~\ref{fig:vis_clip_r50} displays the results on ShanghaiTech, showing density maps predicted by CLIP-EBC (ResNet50).
    \item Fig.~\ref{fig:vis_clip_vit_b} illustrates the results on UCF-QNRF using CLIP-EBC (ViT-B/16).
    \item Fig.~\ref{fig:vis_clip_vit_l} presents the results on NWPU-Crowd with CLIP-EBC (ViT-L/14).
\end{itemize}
In each figure, the images progress from the sparsest to the most congested scenarios from left to right. These visualizations highlight the robustness of our models across a wide spectrum of crowd densities.
%
\end{document}